\documentclass[lettersize,journal]{IEEEtran}
\usepackage{amsmath,amsfonts}
\usepackage{algorithmic}
\usepackage{algorithm}
\usepackage{array}
\usepackage[caption=false,font=normalsize,labelfont=sf,textfont=sf]{subfig}
\usepackage{textcomp}
\usepackage{stfloats}
\usepackage{booktabs}
\usepackage{url}
\usepackage{verbatim}
\usepackage{multirow}
\usepackage{colortbl}
\usepackage{graphicx}
\usepackage{cite}
\usepackage{xcolor}
\usepackage[normalem]{ulem}
\usepackage{soul}
\usepackage{hyperref}
\definecolor{mygray}{gray}{.9}

\begin{document}
	
	\title{Towards Lifelong Scene Graph Generation with Knowledge-ware In-context   Prompt Learning }
	
	\author{Tao He$^{1}$, Tongtong Wu$^2$, Dongyang Zhang$^{1}$, Guiduo Duan$^1$, Ke Qin$^1$, Yuan-Fang Li$^{2}$  \\ 
 $^1$Laboratory of Intelligent Collaborative Computing, University of Electronic Science and Technology of China
	$^2$Department of Data Science and AI, Faculty of Information Technology, Monash University \\
	{\tt\small tao.he01@hotmail.com,wutong8023@gmail.com, dyzhang@uestc.edu.cn, guiduo.duan@uestc.edu.cn, qinke@uestc.edu.cn, yufang.li@monash.edu} 
	}
	
	\markboth{Journal of \LaTeX\ Class Files,~Vol.~14, No.~8, August~2021}%
	{Shell \MakeLowercase{\textit{et al.}}: A Sample Article Using IEEEtran.cls for IEEE Journals}
	

	\maketitle
	
	\begin{abstract}
		
		Scene graph generation (SGG) endeavors to predict visual relationships between pairs of objects within an image. Prevailing SGG methods traditionally assume a one-off learning process for SGG. This conventional paradigm may necessitate repetitive training on all previously observed samples whenever new relationships emerge,  mitigating the risk of forgetting previously acquired knowledge. This work seeks to address this pitfall inherent in a suite of prior relationship predictions. Motivated by the achievements of in-context learning in pretrained language models, our approach imbues the model with the capability to predict relationships and continuously acquire novel knowledge without succumbing to catastrophic forgetting. To achieve this goal, we introduce a novel and pragmatic framework for scene graph generation, namely Lifelong Scene Graph Generation (LSGG), where tasks, such as predicates, unfold in a streaming fashion. In this framework, the model is constrained to exclusive training on the present task, devoid of access to previously encountered training data, except for a limited number of exemplars, but the model is tasked with inferring all predicates it has encountered thus far. Towards LSGG, we propose to present visual content as textual representations as the input tokens for a pretrained language model, e.g., GPT-2, and develop a rehearsal strategy via an in-context prompt based on a novel knowledge-aware prompt retrieval mechanism. Rigorous experiments demonstrate the superiority of our proposed method over state-of-the-art SGG models in the context of LSGG across a diverse array of metrics. Besides, extensive experiments on the two mainstream benchmark datasets, VG and Open-Image(v$_6$), show the superiority of our proposed model to a number of competitive SGG models in terms of continuous learning and conventional settings. Moreover, comprehensive ablation experiments demonstrate the effectiveness of each component in our model.

	\end{abstract}
	
	
	\section{Introduction}
\label{sec:intro}
 
 
The task of scene graph generation (SGG), which entails the detection and localization of visual objects and their relationships within an image, holds a fundamental position in the computer vision community and has garnered significant attention. In the context of SGG, visual concepts such as objects or attributes are systematically presented as a directed scene graph (SG). This representation proves to be of utmost importance in the broader realm of scene understanding \cite{xiao2018unified}  and extends its influence to diverse vision tasks, including image captioning \cite{gu2019unpaired}, image retrieval\cite{krishna2017visual}, and visual question answering \cite{hudson2019gqa}. Formally, a scene graph is articulated as a set of relation triples, denoted as $<$subject, predicate, object$>$.

A substantial body of recent research has made continuous strides in enhancing the performance of SGG~\cite{zellers2018neural,xu2017scene,li2021bipartite,gu2019scene,he2021exploiting,yin2018zoom,kolesnikov2019detecting,hung2020contextual,tang2020unbiased,lyu2022fine,zareian2020bridging,chen2019knowledge,dong2022stacked}. Nevertheless, these methods commonly operate under the assumption of a closed set of predicates, conducting once-and-for-all training on a fixed dataset, such as Visual Genome~\cite{krishna2017visual}. While this conventional training and inference setup is convenient, it proves impractical in realistic applications where new relationships may emerge over time. In practical scenarios, those existing methods necessitate the integration of new data with all previously available data, followed by retraining on the combined dataset. This technique becomes increasingly resource-intensive and time-consuming, particularly as the size of the training set grows substantially. Conversely, training these models solely with new samples presents a challenge known as catastrophic forgetting~\cite{mccloskey1989catastrophic,9349197}. 


 Consequently, an intuitive question emerges: \textit{is it feasible for an SGG model to undergo incremental training with newly arriving data without experiencing forgetting previously acquired knowledge?} In pursuit of an answer to this question, we contemplate a more pragmatic \textit{lifelong learning setting} (alternatively termed continual learning~\cite{9349197}) specific to scene graph generation, referred to as lifelong scene graph generation (LSGG). In this context, an SGG model performs continual updates with newly arriving data. It is notable to highlight that a scene graph comprises object and predicate labels, both of which have the potential to emerge over time. In this study, we narrow the focus of lifelong learning to the latter category, novel predicates, while assuming that all object labels are predetermined through a pretrained object detection network. 
 
 Addressing LSGG has to confront two primary challenges. Firstly, prevalent SGG datasets, such as Visual Genome \cite{krishna2017visual}, manifest a notably long-tailed predicate distribution. Consequently, each task within the LSGG paradigm may exhibit a highly skewed distribution, where certain predicates may be associated with only a few samples. This necessitates endowing the model with a few-shot learning capability for effective learning of the current task. Secondly, since the model lacks access to previously encountered training examples or simply replays with a few-shot examples~\cite{rebuffi2017icarl,rolnick2019experience,de2021continual}, it must not only learn novel relationships using newly arriving data in a few-shot manner but also retain the acquired knowledge from earlier tasks. 



 In addressing the first challenge, we propose predicting visual relationships using pretrained language models, such as GPT-$2$~\cite{radford2019language}, by distilling rich language knowledge from a cross-modality model, e.g., GLIP \cite{zhang2022glipv2}. Recently, pretrained models \cite{brown2020language,devlin2018bert,radford2021learning} have demonstrated considerable success in few-shot and zero-shot scenarios owing to their robust representation capabilities derived from unsupervised training on extensive corpora. A prevalent approach to harnessing these models involves designing or learning \cite{chen2021adaprompt} task-specific prompts for extracting knowledge from them, particularly for downstream tasks such as text generation \cite{li2021pretrained}. However, in SGG, there is no universally pretrained visual relationship model. While the prior work \cite{he2022towards} has proposed pretraining a relation model by leveraging dense captions~\cite{krishna2017visual}, the acquisition of such language corpus proves to be resource-intensive and limits the method's applicability. To circumvent this challenge, we initially extract visual features from a cross-modality model, e.g., GLIP \cite{zhang2022glipv2}. Subsequently, we train an encoder to embed these features as a set of symbolic representations. These embeddings, enriched with linguistic knowledge, then serve as input for a pretrained language model to predict relationships.

The second challenge in LSGG pertains to mitigating forgetting during training. In this regard, L2P \cite{du2022learning} proposed learning multiple prompts to preserve acquired knowledge. However, L2P did not formulate an effective rehearsal strategy for replaying exemplars from the memory buffer. Drawing inspiration from the widely employed in-context learning strategy in natural language processing \cite{chen2021meta,min2022rethinking,lu2021fantastically}, we propose a rehearsal strategy centered around an in-context prompt to retain learned tasks to the greatest extent possible. The fundamental concept behind in-context learning involves incorporating a few examples with ground-truth labels into a prompt, transforming these exemplars into demonstrations \cite{min2022rethinking} or supplements to the conventional prompt \cite{zhou2022learning}. Building upon this notion, we propose the learning of multiple knowledge-aware prompts, each of which retains a few examples for rehearsal. During inference, we employ a knowledge-based retrieval technique to identify the most suitable prompts and their corresponding exemplars. Subsequently, these prompts and exemplars are concatenated to form a comprehensive prompt for the language model to predict relationships.

In summary, our contributions are four-fold:

\begin{itemize}
	 \item We propose a new challenging and practical task, dubbed lifelong scene graph generation (LSGG), aiming to learn to predict predicates in a streaming manner without forgetting the learned knowledge in the past. 
	 
	 \item Towards LSGG, we propose to present visual contents with rich textual symbolic representation via a transformer-based encoder and then design a rehearsal strategy based on an in-context prompt to alleviate forgetting.  To the best of our knowledge, this is the first attempt for LSGG.

     \item  For identifying better prompts and exemplars in the prompt context, we introduce a knowledge-aware prompt retrieval strategy. This strategy proves to be highly advantageous for relation prediction and is particularly effective in mitigating the effects of forgetting.
     
	 \item We conduct extensive experiments over a number of state-of-the-art SGG models to evaluate their performance in the LSGG setting. Results show that our in-context-based prompting method is superior to other models by a large margin, setting a strong baseline for the LSGG task.  Besides, on the task of conventional SGG tasks, our model also exhibits considerable improvements over many SGG models.  
\end{itemize}

	\section{Related Work}

\subsection{Scene Graph Generation}

Scene graph generation \cite{xu2020survey}, also referred to as visual relationship detection \cite{lu2016visual}, is a task focused on detecting and localizing relationships between subject-object pairs within an image. In the early stages of research \cite{lu2016visual,zhang2017visual,xu2017scene,newell2017pixels,dai2017detecting}, the prevailing pipeline involved leveraging an object detection network to extract object features. Subsequently, a feature refinement module was applied to capture contextual cues \cite{zellers2018neural}, external language priors \cite{lu2016visual}, or localization information \cite{inayoshi2020bounding}. The primary research question during this phase often centered on the second step, specifically on how to learn robust representations for relationship classification.
However, with the revelation of a highly skewed predicate distribution in datasets such as Visual Genome \cite{krishna2017visual}, subsequent works \cite{chen2019knowledge,tang2019learning,tang2020unbiased,he2020learning,lyu2022fine} shifted their focus to address the long-tail problem. In this line of research, various techniques were developed, including knowledge embedding \cite{chen2019knowledge}, reweighting or resampling \cite{zellers2018neural}, causal analysis \cite{tang2020unbiased}, fine-grained predicate prediction \cite{lyu2022fine}, and predicate probability distribution based loss (PPDL) \cite{Li_2022_CVPR}. 

However, these methods typically assume that the training data are arriving at once.
In practical scenarios, where the collection of scene graph data is particularly challenging for rare predicates, there is a need for models to continuously learn when new data becomes available without access to previously trained samples. Addressing this gap, we propose a novel task, Lifelong Scene Graph Generation.

\subsection{Prompt-based Learning} 
Prompt learning \cite{liu2021pre} has emerged as a pivotal paradigm in recent years, fueled by the great advancements in pretrained language models (PLMs) such as GPT-3 \cite{brown2020language}. The exploration of prompt-based learning has been particularly pervasive in the realm of natural language processing (NLP), where researchers have harnessed PLMs as expansive knowledge repositories. This approach involves the strategic design of templates or prompts to query PLMs, obviating the need for extensive model retraining. Notable contributions in NLP include the development of adaptive prompts for downstream tasks, such as Adaprompt \cite{chen2021adaprompt}, Prefix-tuning \cite{he2022towards}, Prefix-Tuning \cite{li2021prefix}, and Learning Continuous Prompts \cite{du2022learning}. These efforts underscore the flexibility and efficiency of prompt-based learning in extracting task-specific information from pretrained language models.

The success and versatility of prompt-based learning have transcended the domain of NLP  into computer vision, marked prominently by the advent of Contrastive Language-Image Pre-training (CLIP) \cite{radford2021learning}, which is a large-scale pretrained cross-modality model and has become a linchpin for prompt-based learning in computer vision. In this context, prompt-based learning has demonstrated its efficacy in tasks such as image classification \cite{Li_2022_CVPR} and open-vocabulary object detection \cite{Du_2022_CVPR}. These applications showcase the adaptability of prompt-based learning across diverse domains, underscoring its potential as a versatile methodology for knowledge extraction and task adaptation.

The exploration of prompt-based learning extends beyond the individual model level to the broader context of few-shot learning strategies. The concept of in-context learning has gained prominence in few-shot scenarios for pretrained models in both NLP \cite{lampinen2022can,chen2021meta,min2022rethinking,chiu2021detecting} and computer vision \cite{zhou2022learning}. In-context learning \cite{shin2022effect} involves enriching the input prompt by incorporating few-shot exemplars or demonstrations within the context. This contextual information guides the pretrained model to make more informed predictions without the need for extensive fine-tuning. Such approaches \cite{chen2021meta,kossen2023context} address the challenge of finding optimal prompts for specific queries, enhancing the robustness and adaptability of prompt-based learning strategies.


\subsection {Lifelong learning}

Lifelong learning \cite{9349197} a paradigm that reflects the continuous acquisition of knowledge over time, has garnered significant attention across various domains. In the realm of machine learning and artificial intelligence, lifelong learning has become a pivotal topic, addressing the challenges posed by evolving datasets, dynamic environments, and the need for models to adapt to new information without catastrophic forgetting. Approaches in lifelong learning can be broadly categorized into three main strategies: rehearsal-based methods, parameter isolation methods, and episodic memory methods. Rehearsal-based methods aim to mitigate forgetting by storing and periodically revisiting previously encountered data. Early works such as Elastic Weight Consolidation (EWC) \cite{kirkpatrick2017overcoming} introduced the concept of selectively protecting important parameters based on their significance to previously learned tasks. This strategy was later extended in works like Progressive Neural Networks (PNN) \cite{rusu2016progressive}, which proposed the addition of task-specific neural networks to the model architecture, allowing for the continual integration of new tasks. Parameter isolation methods focus on segregating parameters associated with different tasks to prevent interference during training. One prominent example is the Learning without Forgetting (LwF)  \cite{li2017learning}, where knowledge from previous tasks is distilled into a knowledge base and used to constrain the learning of subsequent tasks. This strategy enables models to leverage previously acquired knowledge while adapting to new tasks. Memory-augmented neural networks, such as Neural Turing Machines (NTM) \cite{graves2014neural}, integrate external memory components to facilitate the storage and retrieval of past experiences. These models enable the effective handling of sequential tasks and provide mechanisms for selective memory access. Beyond these strategies, lifelong learning has also seen the integration of meta-learning techniques. Meta-learning methods, including Model-Agnostic Meta-Learning (MAML) \cite{finn2017model} and Reptile \cite{nichol2018first}, aim to instill a capacity for rapid adaptation to new tasks by exposing models to diverse scenarios during training.
 
 In the context of visual recognition, lifelong learning has been applied to tasks such as object recognition, scene understanding, and image classification.  Paris \textit{et al.} \cite{parisi2017lifelong} employ ensemble learning to adapt to new tasks, while   Hou \textit{et al.} \cite{hou2018lifelong} leverage knowledge distillation for continual adaptation. Visual domain tasks often present additional challenges due to the high dimensionality of image data and the intricate relationships between visual elements.


\section{Preliminaries}\label{sec:ps}
\subsection{Problem Statement}

 Lifelong Scene Graph Generation (LSGG) involves training a visual relationship prediction model from a continuous stream of data, where novel relationship categories may emerge dynamically. Formally, the training data are presented as a sequence of tasks $\mathbb{D}=\{(\mathcal{D}_1, \mathcal{Y}_1), (\mathcal{D}_2, \mathcal{Y}_2), \ldots, (\mathcal{D}_T, \mathcal{Y}_T)\}$, where $\mathcal{D}_{t}=\{D_t^{train},D_t^{val},D_t^{test}\}$ represents the training, validation, and testing sets for the $t$-th task. The complete label set is defined as $\mathbb{Y}=\mathcal{Y}_1 \cup \mathcal{Y}_2 \ldots\cup \mathcal{Y}_T$. For any two label sets, it is assumed that $\mathcal{Y}_i \cap \mathcal{Y}_j=\emptyset$.
 During the $t$-th training stage, the model is trained with $\mathcal{D}_t$ under two configurations: (1) the model cannot access any previously seen data points, denoted as $\mathbb{D}_{1\sim t-1} = \{\mathcal{D}_1, \mathcal{D}_2, \cdots, \mathcal{D}_{t-1}\}$; and (2) the model can only access a few examples from the seen tasks, denoted as $\mathcal{D}^s \in \mathbb{D}_{1\sim t-1}$, where $|\mathcal{D}^s|  \ll |\mathbb{D}_{1\sim t-1}|$. This work focuses on the second setting.
 At the $t$-th inference stage, the model is tasked with predicting the labels of all previous tasks, denoted as $\{\mathcal{Y}_1\cup\mathcal{Y}_2\cup \cdots \cup \mathcal{Y}_t\}$.
  
  In SGG, a scene graph is presented as a suite of relation subject-predicate-object (SPO) triples, i.e,, $(y^s,y^r,y^o)$ where $y^s \in   \mathbb{Y}^o$ and $y^r \in \mathbb{Y}^r$ denote an object  and relationship label set, respectively.   In this work, we mainly consider   $\mathbb{Y}^r$ is given in a streaming manner, that is,   an object class could occur in different tasks, but a relationship class must exclusively emerge in one specific-task.  We could mathematically denote them as  $\mathcal{Y}^r_i \cap \mathcal{Y}^r_{j}=\emptyset$ and $\mathcal{Y}^o_i \cap \mathcal{Y}^o_{j} \ne \emptyset$, where  $i$ and $j$ denote  any two training stages.
 
   \begin{figure*}[ht]
	\centering
	\includegraphics[width=1\linewidth]{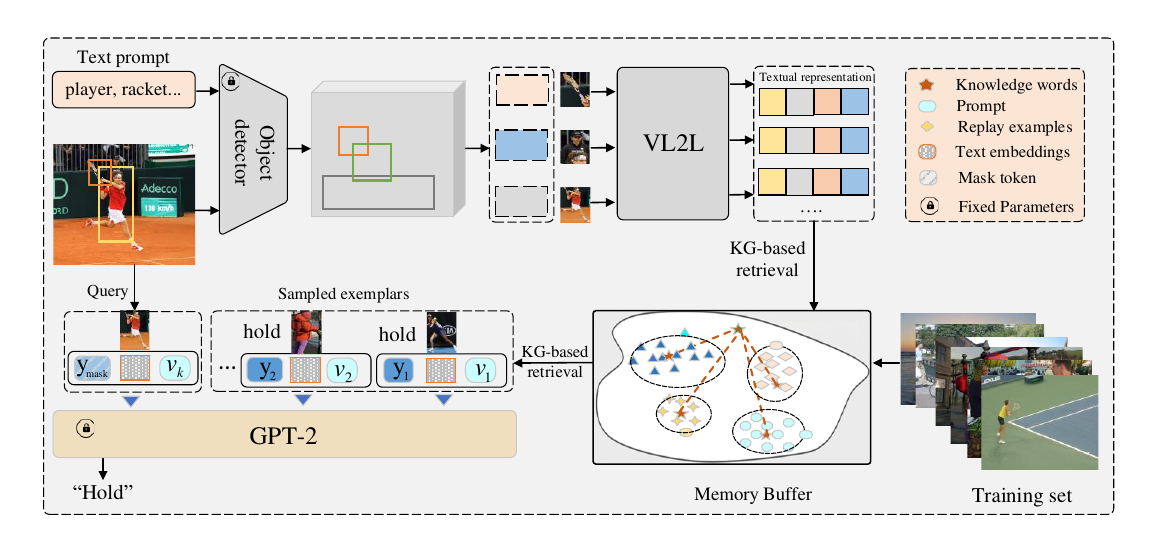}
	\caption{The overall framework of our proposed in-context-based prompt method for life long scene graph generation. There are three main components in our method: (1) visual to textual embedding transformation; (2) knowledge-aware prompts learning;  and (3) in-context exemplars selection for the prompt.   }
	\label{fig:framework}
\end{figure*}

\subsection{Prompt-based learning}
Prompt-based learning aims to exploit a template $\mathcal{T}$ with a $\texttt{[MASK]}$ slot to probe a  pretrained model. For example, in CLIP \cite{radford2021learning},  we can use the embedding of a prompt, e.g.,  ``a photo of \texttt{[class]}'', to classify the object label, rather than learning new classifiers. The challenge is how to find the best prompt given a query. In the above example,  possible prompts could be ``an image of \texttt{[class]}''  or  ``\texttt{[class]} in the photo'', etc.   To solve this problem,  Zhou \textit{et al.} \cite{zhou2022learning} proposed to learn to prompts by finetuning the pretrained model.  Without loss of generality, we could denote the learned prompt as: 
 \begin{equation}
 	\boldsymbol{x}_q=\left[v_{1} ; \cdots ; v_{n^p} ; \boldsymbol{x}_q; \boldsymbol{y}_\texttt{mask}\right]
 	\label{eq:pl}
 \end{equation}
 where $\boldsymbol{x}_q$ is the input query tokens; [;] is the concatenation operation;  $v_i$ denotes the learned soft prompt; $n^p$ is the length of the prompt; and $\boldsymbol{y}_\texttt{mask}$ is the label prediction token. Then, we take $\boldsymbol{x}_i$ as input for a pretrained model and treat the output of $\boldsymbol{y}_\texttt{mask}$ as the predicted result.
 
\subsection{In-context learning}
In-context learning has demonstrated its efficacy in few-shot learning scenarios for pretrained language models \cite{lampinen2022can,chen2021meta,min2022rethinking,chiu2021detecting}. The fundamental notion is to enrich the knowledge embedded in the input prompt by presenting few-shot exemplars complete with ground truth labels in the context. This context guides the pretrained model in making better predictions. Formally, the in-context prompt can be denoted as:
\begin{equation}
	\boldsymbol{x}_q=\left[\boldsymbol{v}; \boldsymbol{x}_1; y_1;\boldsymbol{x}_2; y_2; \cdots;\boldsymbol{x}_{n^e}; y_{n^e};  \boldsymbol{x}_q; y_\texttt{mask} \right]
	\label{eq:in-ctx}
\end{equation}
where $\boldsymbol{v}=[v_{1} ; \cdots ; v_{n^p}]$ represents the prompt in Eq.(\ref{eq:pl}); $(\boldsymbol{x}_j,y_j)$ denotes a pair of input data and the corresponding ground-truth label; and $n^e$ is the number of input examples. In Eq.(\ref{eq:in-ctx}), the context $[ \boldsymbol{x}1; y_1;\boldsymbol{x}2; y_2; \cdots;\boldsymbol{x}{n^e}; y{n^e}]$ can be viewed as a supplement to the prompt in Eq.(\ref{eq:pl}). It provides few-shot exemplars with ground-truth labels, and these exemplars do not introduce extra parameters.


\section{Method}

In Figure \ref{fig:framework}, we illustrate our primary framework for lifelong scene graph generation, which comprises three integral components: the mapping of visual features to symbolic representations, knowledge-aware prompt learning, and in-context exemplar selection. The core idea underlying this framework involves initially presenting visual content as a set of understandable symbolic representations for a pretrained language model. Subsequently, we aim to acquire a set of knowledge-specific prompts, which are stored in a dedicated memory slot. Upon the arrival of a new task, we retrieve the most pertinent prompts along with their associated exemplars and ground-truth labels. These elements are amalgamated to construct a comprehensive in-context prompt, subsequently employed to query the language model.
Specifically, the first component of our framework entails the transformation of visual features into textual tokens or symbolic representations. This transformation is implemented by aligning image features derived from an object detection network with textual space of a language model such as GPT-$2$. For a given query sample, a knowledge-guided retrieval strategy is deployed to identify the most suitable exemplars, which then serve as in-context samples. The selected exemplars, combined with the query sample, are input into a pretrained language model by concatenating them. 


\subsection{Visual features to symbolic representations}

\subsubsection{Visual feature extraction} 
In alignment with prior established SGG models \cite{zellers2018neural,dong2022stacked,lyu2022fine}, our approach involves the extraction of four key features—namely, the context $\boldsymbol{f}^c$, relationship $\boldsymbol{f}^r$, subject $\boldsymbol{f}^s$, and object $\boldsymbol{f}^o$—to represent a subject-object pair comprehensively. Specifically, the context feature is derived from the global image representation, emphasizing a holistic understanding of the entire image. In contrast, the subsequent three features are extracted by an object detection network. Notably, the relationship feature is characterized by the union region of the subject and object within the image, signifying their contextual information in the regional context. 

Traditionally, the widely employed model for object detection in SGG is Faster-RCNN \cite{he2017mask}. However, with the success of pretrained cross-modality models such as RegionCLIP \cite{zhong2022regionclip}, and GLIP \cite{zhang2022glipv2}, cross-modality features from these models have demonstrated greater robustness and generality \cite{zhang2023learning}, along with an open-vocabulary capability. Motivated by this, we choose cross-modality representation as the visual feature. By default, we select GLIP to extract the aforementioned four features. Similar to CLIP, GLIP comprises two encoders designed for images and texts, respectively. Concretely, given an input image, the image encoder of GLIP outputs the region
features $E^v\in \mathbb{R}^{\tilde{N}\times d}$, where $\tilde{N}$ is the number of regions and $d$ is the feature dimension, while the text encoder produces a set of textual embeddings $ E^t \in  \mathbb{R}^{\overline{N}\times d } $, where $\overline{N}$ denotes the number of textual tokens. In the object detection tasks, we could set the object class words as the text input as \cite{zhang2022glipv2}.

\subsubsection{Visual-language to symbolic representations}
After obtaining the cross-modality features $\boldsymbol{f}^* \in E^v$ where $*$ denotes the superscripts $c$, $r$, $s$, and $o$ for brevity, that is the context, relationship, subject, and object feature respectively, we seek to decode them into textual/symbolic representations. To achieve this, we utilize a  standard transformer to map those cross-modality features to textual tokens as \cite{mokady2021clipcap}. Formally, we could denote this as:
\begin{equation}
		 [ {t}^*_1,{t}^*_2, \cdots, {t}^*_{n^*}] = \mathrm{VL2T}( [{\boldsymbol{\phi(f^*)}};  p_1^*, p_2^*, \cdots, p_{n^*}^*])
		 \label{eq:v2l}
\end{equation}
where $\mathrm{VL2T}(.)$ signifies a transformer employed to encode cross-modality features into textual representations; $\phi(.)$ is a project function of MLP to transform the single vector $\boldsymbol{f^*}$ into $l^*$ vectors as the input for  $\mathrm{VL2T}(.)$, i.e., $\boldsymbol{\phi(f^*)} \in \mathbb{R}^{{l^*} \times D}$ in which $D$ is the feature dimension. The parameter $n^*$ denotes the length of the encoded textual tokens. The concatenation operation is denoted by $[;]$. Besides, $\boldsymbol{p}^*=\{p_1^*, p_2^*, \cdots, p_{n^*}^*\}$ represents a set of learnable prompt tokens, and their corresponding hidden output state embeddings are considered as the transformed symbolic representations, i.e., $\boldsymbol{t}^* = \{{t}^*_1, {t}^*_2, \ldots, {t}^*_{n^*}\}$.

\subsection{In-context prompts for LSGG}
 The key challenge in LSGG lies in mitigating the risk of forgetting learned knowledge without recourse to excessive previously trained instances. A na\"ive  prompt-based way method (e.g. \cite{Du_2022_CVPR}) involves inputting tokenized representations, i.e., ${\boldsymbol{t}}^c$, ${\boldsymbol{t}}^r$, ${\boldsymbol{t}}^s$ and ${\boldsymbol{t}}^o$, into a pre-trained language model (e.g., GPT-$2$) as in Eq.(\ref{eq:pl}), i.e.,
 \begin{equation}
 \boldsymbol{x}_q = [ \boldsymbol{v}, \boldsymbol{t}^c, {\boldsymbol{t}}^r, {\boldsymbol{t}}^s, {\boldsymbol{t}}^o, \boldsymbol{y}_{\texttt{mask}} ]
 \label{eq:l2p}
 \end{equation}
Nevertheless, in practice, this approach gives rise to pronounced issues of catastrophic forgetting. We conjecture that this occurrence can be attributed to two discernible reasons: firstly, the extra knowledge reliant on the single prompt $\boldsymbol{v}$ may prove effective for the current task, yet lack adaptability when confronted with a new task; secondly, the contextual information, encompassing solely the global image cues $\boldsymbol{t}^c$, is inherently limited for a language model.
To mitigate the aforementioned issues, we propose learning multiple adaptive knowledge-aware prompts, each of which stores
a  few  representative exemplars, serving as the in-context
in Eq. (\ref{eq:in-ctx}).
 
 \subsubsection{Knowledge-aware prompts learning}
 A main challenge within in-context learning pertains to the selection of suitable examples for contextualization \cite{lampinen2022can,chen2021meta,min2022rethinking}. In Equation (\ref{eq:in-ctx}), prevailing in-context methods \cite{lampinen2022can,min2022rethinking} commonly employ a uniform prompt denoted as $\boldsymbol{v}$ across all contextual exemplars. However, we posit that such an approach proves ineffectual for LSGG. As illustrated by the findings in L2P \cite{wang2022learning}, relying on a singular prompt fails to adequately preserve acquired knowledge. Consequently, we contend that diverse exemplars may be associated with distinct prompts, each representing a distinct type of knowledge. For instance, in SGG, where the VG dataset features multiple predicate types such as localization and human actions, we advocate the utilization of disparate prompts, namely knowledge-aware prompts, for each specific type of knowledge.
 

To that end, we define our knowledge-aware prompts as a collection of triples denoted as $\Phi=\{(\boldsymbol{v}_i \in \mathbb{R}^{{n^p}\times d},\boldsymbol{k}_i \in \mathbb{R}^{512}, \boldsymbol{E}_i)\}_{i=1}^{n^t}$, where $n^t$ represents the number of prompts, $\boldsymbol{k}_i$ signifies a knowledge word, and $\boldsymbol{E}_i=\{e_j\}_{j=0}^{n^e}$ is a set where each element is defined as $\boldsymbol{e}_j=(\boldsymbol{f}^c_j, \boldsymbol{f}^s_j,{\boldsymbol{f}}^o_j, {\boldsymbol{f}}^r_j, y_j)$, corresponding to a seen instance with the ground truth. To elaborate, when presented with a query $e_q$, the initial step involves utilizing the global context feature $\boldsymbol{f}^c_q$ to identify the top-$K$ most similar $\boldsymbol{k}$ through cosine similarity measurements across the set $\Phi$. This retrieval process can be articulated as follows:
\begin{equation}
  \boldsymbol{K}_* = \underset{1\le i \le |\Phi|}{\mathrm{TopK}}   (\mathrm{cosine}(\boldsymbol{f}^c_q, \boldsymbol{k}_i))
  \label{eq:topk}
\end{equation}
 where $\mathrm{TopK}(.)$ denotes a retrieval function designed to yield the top $K$ closest knowledge words. Notably,  $\boldsymbol{k}_i$ undergoes a random initialization in its nascent stage. After the identification of the most similar knowledge words, the corresponding prompts of $\boldsymbol{K}_*$ are selected as the input prompts. 
 
 As for the exemplar, different from the knowledge word retrieval, we employ the regional context representation, i.e., the relationship feature ${\boldsymbol{f}}^r_q$, to retrieve the optimal exemplar within each $\boldsymbol{E}_i$ through a cosine similarity search, defined as:
  \begin{equation}
 	\boldsymbol{e}^{in} = \underset{1\le i \le |\boldsymbol{E}_k|}{\mathrm{TopOne}}   (\mathrm{cosine}({\boldsymbol{f}}^r_q, {\boldsymbol{f}}_i^r))
 	\label{eq:top1}
 \end{equation}
 where $\mathrm{TopOne}(.)$ is  a  function to return the closest exemplar in $\boldsymbol{E}_i$. 
 
 \subsubsection{In-context prompt of relation prediction}
 
So far, we have the top $K$ most similar prompts  $\boldsymbol{K}_*$ along with their corresponding exemplars $\{\boldsymbol{e}_i^{in}\}_{i=1}^{k}$, complete with  their labels $\{\boldsymbol{y}_i\}_{i=1}^{k}$.
 Then, we  could define our in-context prompt as:
 \begin{equation}
 	\boldsymbol{x}_q=\left[\boldsymbol{v}_k; \boldsymbol{e}_k^{in}; \boldsymbol{y}_k;  \cdots; \boldsymbol{v}_{2}, \boldsymbol{e}_{2}^{in}; \boldsymbol{y}_{2}; \boldsymbol{v}_{1}, \boldsymbol{e}_q, \boldsymbol{y}_\texttt{mask} \right]
 	\label{eq:lsgg}
 \end{equation}
Where $\boldsymbol{v}_k \cdots \boldsymbol{v}_1$ represent the retrieved $K$ prompts arranged in ascending order of similarity, as \cite{min2022rethinking}.   
In practice,   $\boldsymbol{e}^{in}=[\boldsymbol{t}^c;\boldsymbol{t}^r;{\boldsymbol{t}}^s,{\boldsymbol{t}}^o]$ is the concatenation of image context, relation region, subject, object tokens, which are transformed by Eq.( \ref{eq:v2l}).
Then, we feed $\boldsymbol{x}_q$ into a pretrained language model as:
\begin{equation}
	 [\boldsymbol{p}_1, \boldsymbol{p}_2, \ldots \boldsymbol{p}_{|\boldsymbol{x}_q|}] = \mathrm{LM}_{\theta} (\boldsymbol{x}_q)
	 \label{eq:lmsgg}
\end{equation}
where $\mathrm{LM}_{\theta}(.)$ denotes a pretrained language model (e.g., GPT-$2$), $\theta$ denotes the parameters of the language model, and $\boldsymbol{p}_i$ is a prediction score distribution. It is important to note that, in practical scenarios where relationship words may comprise multiple words (e.g., "sitting on"), $\boldsymbol{y}_\texttt{mask} \in \mathbb{R}^{n^r \times d }$ comprises $n^r$ learnable embeddings, where $n^r$ denotes the maximum length encompassing all potential predicate words.

 \subsection{Training and Inference} 
 
\subsubsection{Training}

In summary, our framework comprises four primary modules: the cross-modality feature extraction network, GLIP; the transformation from visual features to symbolic representations,   $\mathrm{V2L(.)}$ in Eq. (\ref{eq:v2l}); the learnable knowledge words stored in ${\Phi}$; and the relation predictor, $\mathrm{LM(.)}$ in Eq.(\ref{eq:lmsgg}). It is noteworthy that, during the training phase, we froze the parameters of both the feature extraction network, GLIP, and the language model, but only learn the $\mathrm{V2L(.)}$, the prompts and knowledge words.

 
 Intuitively, the learning of knowledge words  is equal to optimize the distance between the query $\boldsymbol{f}_q^c$ and  the selected knowledge words $\boldsymbol{K}_*$, that is:
 \begin{equation}
 \mathcal{L}_2 = \sum_{i=1}^{|\boldsymbol{K}_*|} \mathrm{cosine}(\boldsymbol{f}_q^c, \boldsymbol{k}_i)
 \end{equation}
 
 Regarding the predicate classification, we  use the  standard  cross-entropy loss to classify them as:
 \begin{equation}
 	\mathcal{L}_3 \!\!=\!\! -\!\!\!  \sum_{j=1}^{j=n^r} \!\! \log p\left( {y}_{m+j}^r \!\!\mid \!\boldsymbol{p}_1, \ldots,  \boldsymbol{p}_{m+1}, \ldots, \boldsymbol{p}_{m+j-1} \right)
 \end{equation}
 where $m$ is the number of length of the prefix in $\boldsymbol{x}_q$, i.e., $ \left|\boldsymbol{v}_k; \boldsymbol{e}_k^*; \boldsymbol{y}_k;  \cdots; \boldsymbol{v}_{2}, \boldsymbol{e}_{2}^*; \boldsymbol{y}_{2}; \boldsymbol{v}_{1}, \boldsymbol{e}_q \right|$; ${y}_{m+j}^r$ denotes the $j$-th predicate word in $\boldsymbol{y}_{mask}$; $\boldsymbol{p}_i$ is an output probability for the $i$-th  example    and each $|p_i|$ is the vocabulary size of the language model; and $\boldsymbol{y}_j^r$ denotes a ground-truth predicate.  It is worth  noting that during training, the examples in buffer memory also require to be replayed by $\mathcal{L}_3$, serving as rehearsal loss\cite{9349197,abs-1904-07734}. Hence, the overall loss 
 is:
 \begin{equation}
 	\mathcal{L} =  \alpha\mathcal{L}_1 +  \lambda\mathcal{L}_2 +   \mathcal{L}_3
 \end{equation}
 where $\alpha$ and $\lambda$ are two hyperparameters to weight the terms.
 
 \subsubsection{Inference}
 we employ Eq. (\ref{eq:topk}) to retrieve the top $K$ similar knowledge entries from the buffer memory, which is populated after training. Subsequently, we apply Eq. (\ref{eq:top1}) to identify the most similar exemplars. These retrieved knowledge entries and exemplars are then concatenated to form an extended input sequence denoted as $\boldsymbol{x}_q$ in Eq. (\ref{eq:lsgg}), which is utilized as input for the language model. 

  \section{Experiments}
 In this section, we will first elaborate on our experimental dataset, evaluation setting and metrics. Then, we will report our model's    quantitative  performance on LSGG,  compared with   the state-of-the-art methods. Last, a comprehensive ablation study and qualitative results are presented.  

 \subsection{Experiment Settings}
 
 \subsubsection{Dataset}
 
  \noindent \textbf{Visual Genome} (VG) is the mainstream benchmark dataset for SGG. Following previous works~\cite{zellers2018neural,tang2020unbiased,xu2017scene}, we use the pre-processed VG with $150$ object classes and $50$ predicates~\cite{xu2017scene}. VG consists of 108k images, of which $57,723$ images are used for training and $26,443$ for testing. Additionally, $5,000$ images make up the validation set.
\noindent \textbf{Open-Image}(v$_6$): consists of $301$ object
categories and $31$ predicate categories. Following the split of \cite{han2021image}, the training set has 
$126,368$ images while the validation and test sets contain $1,813$ and $5,322$ images respectively. 

\subsubsection{Evaluation settings of LSGG} In the context of LSGG, we consider training data are arriving in a sequential manner. To emulate this scenario, we partition the predicate words of VG into $5$ sub-sequences. For instance, on the VG dataset, each task may contain $10$ predicates. Regarding task division, we employ two distinct splitting strategies:  Random Splitting and Frequency-Based Splitting, but in default, we use the random splitting strategy. Specifically, the former randomly divides the predicates into five tasks,  commonly adopted in mainstream continuous learning studies \cite{chaudhry2019tiny,buzzega2020dark,rolnick2019experience}. The latter approach aligns with the frequencies of predicates to divide those predicates, adhering to a long-tail distribution observed in the training set. The rationale behind this strategy is rooted in the belief that more common predicates are relatively easier to acquire, whereas the rare ones pose a greater challenge and are anticipated to arrive later in the training sequence. During any training stage, the model lacks access to seen examples or is restricted to a limited number of examples from prior training stages. After the completion of each training stage, model evaluation is conducted concerning all arrived predicates.  Our experimental evaluations are reported across three sub-tasks in SGG: Predicate Classification (\textsc{PredCls}), Scene Graph Classification (\textsc{SGCls}), and Scene Graph Detection (\textsc{SGDet}).


\subsubsection{Evaluation Metrics}

We mainly report the results on two types of metrics:  the unbiased metric mean Recall@K ($\mathbf{{mR@K}}$)  and the conventional metric $\mathbf{{R@K}}$. It is worth noting that the latter one does not reflect a model's true performance on tail relations~\cite{chen2019knowledge,tang2020unbiased}. Hence, following \cite{zheng2023prototype}, we also report the results on the average of R@K and mR@K, denoted as M@K. Regarding continuous learning, we report the model's performance to prevent forgetting on the metric of \textbf{Forgetting Measure}~\cite{chaudhry2018efficient,chaudhry2018riemannian,chaudhry2019tiny}.    Besides, for Open-Image, following the settings of \cite{han2021image}, we report four met-
rics: R@$50$, weighed mean Average Precision (wmAP$_{rel}$) and weighed mean Average
Precision (wmAP$_{phr}$) of triples, and score$_{wtd}$. The score$_{wtd}$ is calculated as: scorewtd = $0.2$ $\times$ R@$50$ + $0.4$ $\times$ wmAP$_{rel}$ +
$0.4$ $\times$ wmAP$_{phr}$.  
  \begin{table*}[!ht]
 	\centering
  \caption{The  comparison results of LSGG with the buffer size set to   $2,000$ after the five training stages. We report the results on  \textsc{PredCls}, \textsc{SGCls}, and \textsc{SGDet} tasks of VG$150$ with respect to mR@$50$/$100$ and R@$50$/$100$.   }
 \resizebox{2\columnwidth}{!}
 {
 	\begin{tabular}{l|ccc|ccc|ccc}
 		\toprule
 		\multirow{2}{*}{Methods}   & \multicolumn{3}{c}{{Predicate Classification}} & \multicolumn{3}{c}{Scene Graph Classification} & \multicolumn{3}{c}{Scene Graph Detection} \\
 		&mR@50/100     & R@50/100   & M@50/100 & mR@50/100    & R@50/100 & M@50/100 & mR@50/100    & R@50/100  & M@50/100 \\\midrule
 		IMP\cite{xu2017scene} &       ~~7.7 / ~~8.5      &     45.7 / 47.2    & 26.7 / 27.9     &     5.0  / 5.7          &  25.6 / 27.0    &  15.3/ 16.4   &   2.6 / 3.4           &  16.7 / 18.8     & 9.7 / 11.1     \\
 		Motifs\cite{zellers2018neural}  &       ~~9.6 / 11.2             &   50.2 / 51.9  & 29.9 / 31.6 &  6.4 / 7.2     &      28.3 / 30.9  & 17.4/ 19.1      &   3.5 / 4.2                       &   18.8  / 20.9    & 11.2 / 12.6    \\
 		VCTree\cite{tang2019learning}  &     10.8  / 12.6         &   51.7 / 53.1   & 31.3 / 32.9      &     6.6 / 7.8         &  27.9 / 29.2    & 17.3 / 18.5      &    3.4 / 4.5          &     17.2 / 19.4     & 10.3 / 12.0   \\  
 		\multirow{1}{*}{TDE\cite{tang2020unbiased}} 
 		&       12.2 / 13.5          &      35.4 / 37.7   & 23.8 / 25.6    &     7.0 / 8.1         &    20.1   / 22.4   & 13.6 / 15.6  &     3.3 / 4.0         &      15.7 / 17.7 & 9.5 / 10.9    \\ 
 		\multirow{1}{*}{SHA\cite{dong2022stacked}} 
 		&         15.2 / 16.9        &   38.2 / 41.1     & 27.0 / 29.3     &   8.3 / 9.1         &   22.5 / 23.8    & 15.4 / 16.5     &    3.7 / 4.4          &        17.8 / 19.2 & 10.8 / 11.8  \\  
 		\multirow{1}{*}{SQUAT\cite{jung2023devil}} 
 		&    14.1 / 15.7   &   50.6 / 52.4    & 32.4 / 34.1  &       7.7 / 8.6       &     26.1 / 28.0   & 16.9 / 18.3   &    4.0 / 4.6         & 19.5 / 21.4    & 11.8 / 13.0       \\ 
   
 		\multirow{1}{*}{Ov-SGG~\cite{he2022towards}} 

 		&         15.3 / 17.0       &  51.2 / 52.7   & 33.3 / 34.9        &       8.0 / 9.3      &   29.4 / 31.8   & 18.7 / 20.4      &    -- / --          &   -- / --  & -- / --      \\
      PE-Net \cite{zheng2023prototype}  & 15.6 / 17.4 & 52.1 / 54.8 & 33.9 / 36.0 & 9.1 / 10.8 & 31.4 / 32.9 & 20.4 / 22.1 & 4.2 / 5.3 &  21.2 / 23.8 & 12.7 / 14.6\\
   {VS$^3$}\cite{zhang2023learning} &     16.2 /  17.1        &    52.3 / 54.2  & 34.3 / 35.7         &  8.8 / 10.2         &  31.2 / 32.5    & 20.0 / 21.4      &    4.5 / 5.1          &  20.1 / 22.4 & 12.3 / 13.8\\
 		\midrule   
 		
 		\textbf{ICSGG}$_{s}$&    {14.6} /  {15.7}       &  51.0 / 52.7    & 32.8 / 34.2         &   8.6 / 9.5        &   30.2 / 31.7   & 19.4 / 20.8      &  3.8 / 4.5            &   19.4 / 20.3 & 11.6 / 12.4   \\
 		\textbf{ICSGG}$_{m}$&        {15.4} /  {16.6}      &   52.9 / 54.6  & 34.2 / 35.6  &     9.0 / 10.1        &     31.6  / 33.0  & 20.3 / 21.6  &      4.0 / 4.8       &  20.0 / 21.7  & 12.0 / 13.5  \\ 
 		\textbf{ICSGG} &         \textbf{18.6} / \textbf{20.3}        &   \textbf{54.1} / \textbf{56.4}     & \textbf{36.4} / \textbf{38.4}     &    \textbf{9.7} / \textbf{11.4}         &  \textbf{33.1} / \textbf{34.8} & \textbf{21.4} / \textbf{23.1}         &     \textbf{4.9} / \textbf{5.6}         &   \textbf{22.7} / \textbf{24.5}  & \textbf{13.8} / \textbf{15.1} \\
 		\bottomrule     
 	\end{tabular}
 	}
 	
 	\label{tb:lsgg2k}
 	
 \end{table*}

\subsubsection{Baseline methods} 

We select a number of representative SOTA models as our baselines: IMP~\cite{xu2017scene}, Motifs~\cite{zellers2018neural}, VCTree~\cite{tang2019learning}, TDE~\cite{tang2020unbiased}, 
   SHA~\cite{dong2022stacked}, Ov-SGG~\cite{he2022towards}, VS$^3$\cite{zhang2023learning}, and PE-Net \cite{zheng2023prototype}. Note that since TDE and SHA are model-agnostic, for a fair comparison, we choose VCTree as the their base model. Our model is dubbed as ICSGG.

 \subsection{Implementation Details}
Following the setup in VS$^3$ \cite{zhang2023learning}, we employ GLIP (i.e., the GLIP-T and larger GLIP-L encoders) as our object detection network, and their parameters remain frozen during training. Consistent with \cite{zhang2023learning}, we retain the top $36$ object detection results per image for scene graph detection.
The $\mathrm{VL2T}$ comprises an $8$-layer transformer with multi-head attention components. We represent the context, relationship, subject, and object using $4$, $4$, $2$, and $2$ tokens, respectively. The input token dimension for the language model is set to $768$. Two hyperparameters, $\alpha$ and $\lambda$, are set to $0.2$ and $0.5$, respectively.
For our multiple prompts, we default to $100$ prompts, each storing $20$ examples, akin to the buffer size setting in \cite{rebuffi2017icarl}. Each prompt consists of $8$ tokens, with $n^p$ set to $8$. We utilize GPT-$2$ ($774$M) as our language model with a vocabulary size of $50,257$.
Additionally, we experiment with other language models, such as GPT-$2$ small ($117$M) and GPT-$2$ Medium ($345$M). All experiments are conducted on $8$ Nvidia $4090$Ti GPUs using the AdamW optimizer with a learning rate of $0.002$ and a weight decay of $10^{-4}$.
Our model implementation is based on the released code of TDE \footnote{https://github.com/KaihuaTang/Scene-Graph-Benchmark.pytorch} and the open-source platform \footnote{https://huggingface.co}.
  
 \subsection{Main Results of LSGG and SGG}
 We first report the overall performance after the five training stages with the buffer size set to   $2,000$.  All models use the same object detection network  GLIP \cite{zhang2022glipv2} to extract region features. For the other baseline models, we randomly sample examples to store in the memory buffer.   

\subsubsection{Overall results of LSGG} Table \ref{tb:lsgg2k} shows the comparison results of LSGG after the five training stages on the VG dataset. For the baseline models, following the incremental setting of \cite{rebuffi2017icarl}, we store $2,000/M$ examples dynamically for each seen predicate, where $M$ is the number of seen predicates and increases with the training stage. Note that we further report other variant our models, i.e., ICSGG$_s$ and ICSGG$_m$, using the small size GPT-$2$ ($117$ MB) and medium size GPT-$2$ ($345$ MB), respectively.
It is worth noting that VS$^3$, originally conceived as a weakly supervised scene graph generation model, is evaluated in our comparison based on its fully supervised version.  
 \begin{table}[t]
	\centering
 	\caption{ 
The  comparison results of LSGG on the Open-Image(v$_6$) dataset with the buffer size set to   $2,000$ after the five training stages. 
}
		\begin{tabular}{lcccc}
			\toprule
			{Models}& Recall@50    &  wmAP$_{rel}$   &     wmAP$_{phr}$   & score$_{wtd}$ \\  \midrule
			Motifs \cite{zellers2018neural}    & 58.27 &  21.59 & 23.86 &   29.83  \\
			VCtree\cite{tang2019learning}     &  59.32   &   22.08   & 24.15  &     30.35    \\
		SQUAT~\cite{jung2023devil}        &  60.13   &   23.05   & 23.12  &    30.49    \\
   SHA~\cite{dong2022stacked}        &  61.41   &   23.14   & 24.28  &    31.25   \\ 
   PE-Net~\cite{zheng2023prototype}        &  61.35   &   23.80   & 24.17  &    31.86     \\
			OV-SGG~\cite{he2022towards}        &  61.21  &   24.12   & 25.36  &  32.03     \\ 
   	
   VS$^3$~\cite{zhang2023learning}        &  62.08   &   24.48   & 25.69  &    32.48   \\ 
			\midrule
   \textbf{ICSGG$_s$}       &  62.73   &   24.70  & 26.21  &  29.97      \\ 
   \textbf{ICSGG$_m$}      &  64.82   &   25.41   & 27.38  &    34.08     \\ 
			\textbf{ICSGG}                      &  \textbf{65.24}   &   \textbf{25.97}   & \textbf{28.39}   &   \textbf{34.79}    \\
			\bottomrule
		\end{tabular} 
	\label{tab:oi}
 
\end{table}

 For the results, we could observe  that  many SOTA methods, such as  SHA and PE-Net, commonly underperform on the new task of LSGG, especially on the metric of mR@K, although they have shown competitive performance on the conventional SGG setting. Interestingly, VS$^3$ gains competitive results across many metrics, and beat  all the other  prior models in the table. We posit that the primary contributing factor to this observation lies in the substantial inclusion of a large language supervision in VS$^3$. This incorporation significantly enhances the model's grasp of visual-language knowledge, thereby mitigating the extent of forgetting on the acquired knowledge.
 Our ICSGG gains the best results over all terms. For example,  on the task of \textsc{PredCls}, ICSGG exceeds the second best model VS$^3$ by  $2.8$ on average in terms of mR@$50$/$100$. Similarly, the same case can be found on the other baseline models, e.g., SHA~\cite{dong2022stacked} despite of showing competitive SOTA performance on the conventional SGG. Table \ref{tab:oi} presents the results on the Open-Image (v$_6$) dataset. Similar to the observations on the VG dataset, our  ICSGG  model attains the most favorable performance across all metrics, exhibiting an approximately $2$-point superiority over the second-best model VG$^3$ on average. This underscores the efficacy of our in-context-based knowledge-aware prompt learning approach in the context of lifelong scene graph generation. In the ablation section, we will delve into a detailed evaluation of the effectiveness of each component within our model.

 
 \subsubsection{Results of forgetting} An importance   metric  of lifelong learning is Forgetting Measure(FM)~\cite{chaudhry2018riemannian}, which gauges how effectively a model retains previously acquired knowledge..  Notably, a higher FM value indicates a greater extent of forgetting, signifying a worse performance on tasks from previous stages.   We use mR@K as the base measurement to calculate FM. Table \ref{tb:fm} shows the results of overall FM after five training stages. It is evident that conventional  SGG  models, such as Motifs and Ov-SGG, and even recent state-of-the-art methods like VS$^3$, exhibit more pronounced forgetting issues compared to our model. However, ICSGG demonstrates an average FM value that is $1.3$ points lower than the second-best model, VS$^3$, across all terms.

 \subsubsection{Results after each training stage} 
\begin{figure}[t]
 	\centering
 	\includegraphics[width=1\linewidth]{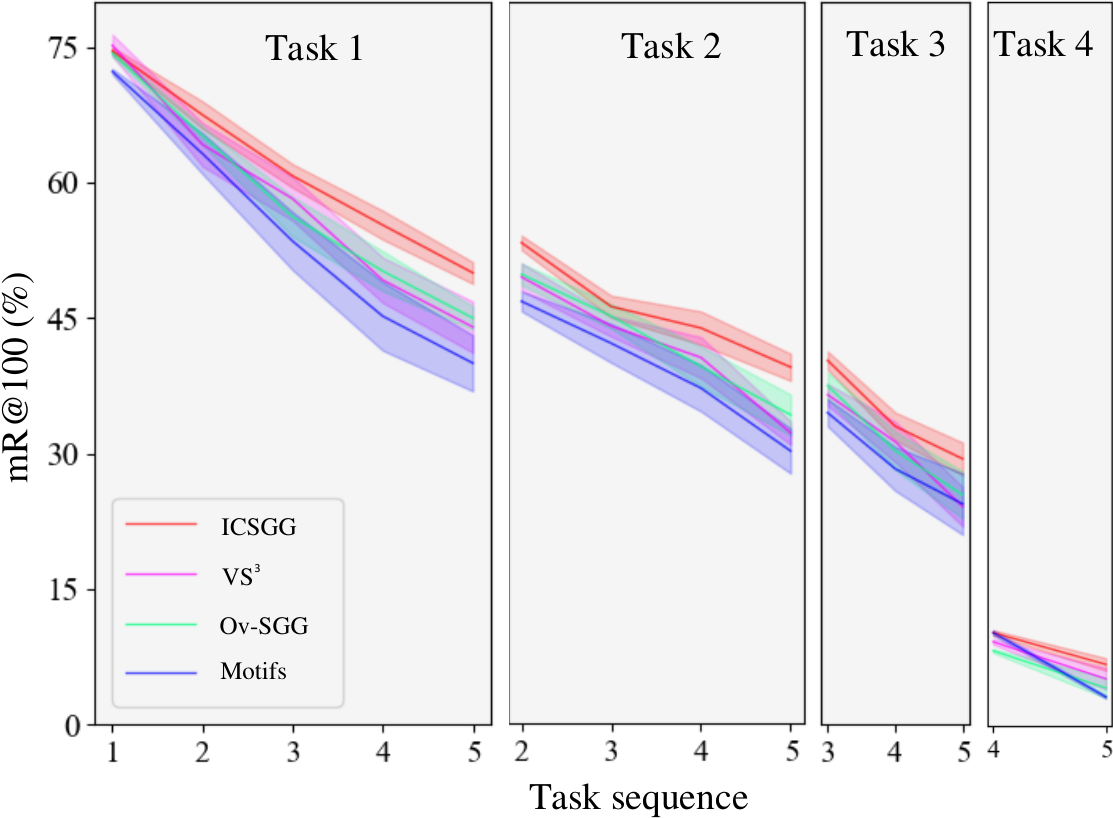}
 	  \caption{ Results of four tasks at  different training stages on the \textsc{PredCls} of mR@$100$ on VG.  }
 	\label{fig:sts}
 \end{figure}

We also present the results of \textsc{PredCls} from task$_1$ to task$_4$ in terms of mR@$100$ after each training stage on the VG dataset, as illustrated in Figure \ref{fig:sts}. The outcomes reveal that, while several baseline models such as Motifs and VS$^3$ exhibit similar performance initially, they manifest severe forgetting issues as training progresses. For instance, in task$_1$, VS$^3$ and ICSGG display comparable performance initially, but after five training stages, ICSGG demonstrates an approximately $5$-point improvement over VS$^3$. These findings provide additional confirmation of our model's proficiency in mitigating catastrophic forgetting.
\subsubsection{Results of SGG}
For a fair comparison with other SGG models, we also present results using the conventional training scheme, assuming all predicates arrive simultaneously. Table \ref{tab:fs} illustrates the outcomes on the VG dataset, utilizing the unbiased metric of mR@K. It is important to note that all compared models utilize the same object detection results from GLIP and region features for equitable comparison while excluding reweighting \cite{cui2019class} or resampling \cite{tang2020unbiased} strategies. Our memory buffer is set to $2,000$.
From the results, it is evident that our  ICSGG  continues to outperform other SGG models, with the exception of SQUAT on the mR@$100$ metric for \textsc{PredCls}. This underscores the effectiveness of our in-context prompt learning strategy not only in continuous learning but also in its ability to mitigate biases.

 \begin{table}[t]
 	\centering
   \caption{The results on the metric of Forgetting Measure. \textbf{Lower is better}. We set buffer memory to $2,000$ for all methods.  }
 \label{tb:fm}
 	\begin{tabular}{l|ccc}
 		\toprule
 		\multirow{2}{*}{Methods} & \multicolumn{1}{c}{\textsc{PredCls}} & \multicolumn{1}{c}{\textsc{SGCls}} & \textsc{SGDet}     \\
 		& FM@50/100                   & FM@50/100                & FM@50/100 \\\midrule
 		Motifs\cite{zellers2018neural}&         19.2 / 21.3 &  11.2 / 12.9  &   7.1 / 8.2       \\
 		VCTree\cite{tang2019learning}&         18.1 / 20.6 &  10.5 / 12.0  &   6.4 / 7.4       \\
 		
 		SQUAT\cite{jung2023devil}   &           17.3 / 19.5 &  ~~9.8 / 11.3 &   7.4 / 8.1    \\
   PE-Net\cite{zhang2023learning}&            16.5 / 18.4 &  ~~9.2 / 10.4 &   6.3 / 7.0       \\
 		Ov-SGG\cite{he2022towards}&         16.0 / 17.7 &  ~~8.5 / ~~9.4&   6.2 / 6.9       \\ 
   	  VS$^3$ \cite{zhang2023learning}   &         15.1 / 17.2 &  ~~8.0 / ~~9.2&   5.9 / 6.6       \\ 
     \midrule
 		\textbf{ICSGG}$_s$& 15.4 / 17.0  &  ~~8.2 / ~~9.1  &   6.1 / 7.3            \\ 
 		\textbf{ICSGG}$_m$& 14.6 / 16.3  &  ~~7.4 / ~~8.7  &   5.6 / 6.7            \\  
 		\textbf{ICSGG}& \textbf{13.5} / \textbf{15.3} &  ~~\textbf{6.4} / ~~\textbf{8.0}&   \textbf{5.0} / \textbf{5.8}            \\ \bottomrule
 	\end{tabular}
 \end{table}
\begin{table}[t]
	\centering
	\caption{The comparison results of fully supervised SGG on the VG dataset in terms of mR@K. $\dagger$ means that the results are reproduced by their released code. }
	\label{tab:fs}
 {
		\begin{tabular}{l|ccc}
			\toprule
			   \multirow{2.5}{*}{Method} & \textsc{Predcls} & SGCls & \textsc{SGDet} \\   
			& mR@50/100 & mR@50/100 & mR@50/100 \\\cmidrule{1-4} 
			IMP~\cite{xu2017scene} &  11.0 / 11.8  & 5.6 / 5.9  &  3.7 / 4.8 \\ 
			Motifs \cite{zellers2018neural}& 12.7 / 15.8 & 7.6 / 8.8  &  5.9 / 6.7 \\ 
			VCTree \cite{tang2019learning} &  {14.2 / 16.5}&   8.2 / 9.6 &   6.3 / 7.1 \\  
			SHA $\dagger$ \cite{dong2022stacked} & {18.9 / 20.8} &  10.9 / 11.6  & { 7.8 / 9.0} \\ 
			SQUAT$\dagger$ \cite{jung2023devil} & {25.5 /  \textbf{28.3}}  & {17.7 / 19.2} & {13.4 / 15.7} \\
		  Ov-SGG \cite{he2022towards}& 24.3 / 26.3 &  12.5 / 15.3  & 10.5 / 12.8 \\
			 VS$^3$ \cite{zhang2023learning} &  21.4 / 24.9 &   14.7 / 17.1 &  11.5 / 13.9\\ \midrule
    
    	 \textbf{ICSGG$_s$} &  23.4 / 25.2 &   13.6 / 15.5 &  11.5 / 12.4\\
      	 \textbf{ICSGG$_m$} &  24.5 / 26.9 &   15.2 / 17.0 &  12.3 / 13.3\\
        	 \textbf{ICSGG} &  \textbf{25.7} / {27.8} &   \textbf{17.8} / \textbf{20.0} &  \textbf{14.5} / \textbf{16.2} \\\midrule
		\end{tabular}
	}
	
\end{table}
 
 \subsection{Ablation Study}
 In this section, we will  test   effectiveness of main components in our model. Specifically, there are four main techniques:   (1)  in-context exemplar selection and ordering;  (2) the size of the memory buffer;    (3) finetuning the language model;  (4) the length of  prompts; and (5) the task splitting strategy. When conducting    experiments, we only modify the ablated component but keep the other setting to the best. The results on the VG and Open-Image(v$_6$) are shown  in Table \ref{tb:ab} and Table \ref{tab:ab_oi}, respectively.  
   \begin{figure*}[!ht]
 	\centering
  
 	\includegraphics[width=1\linewidth]{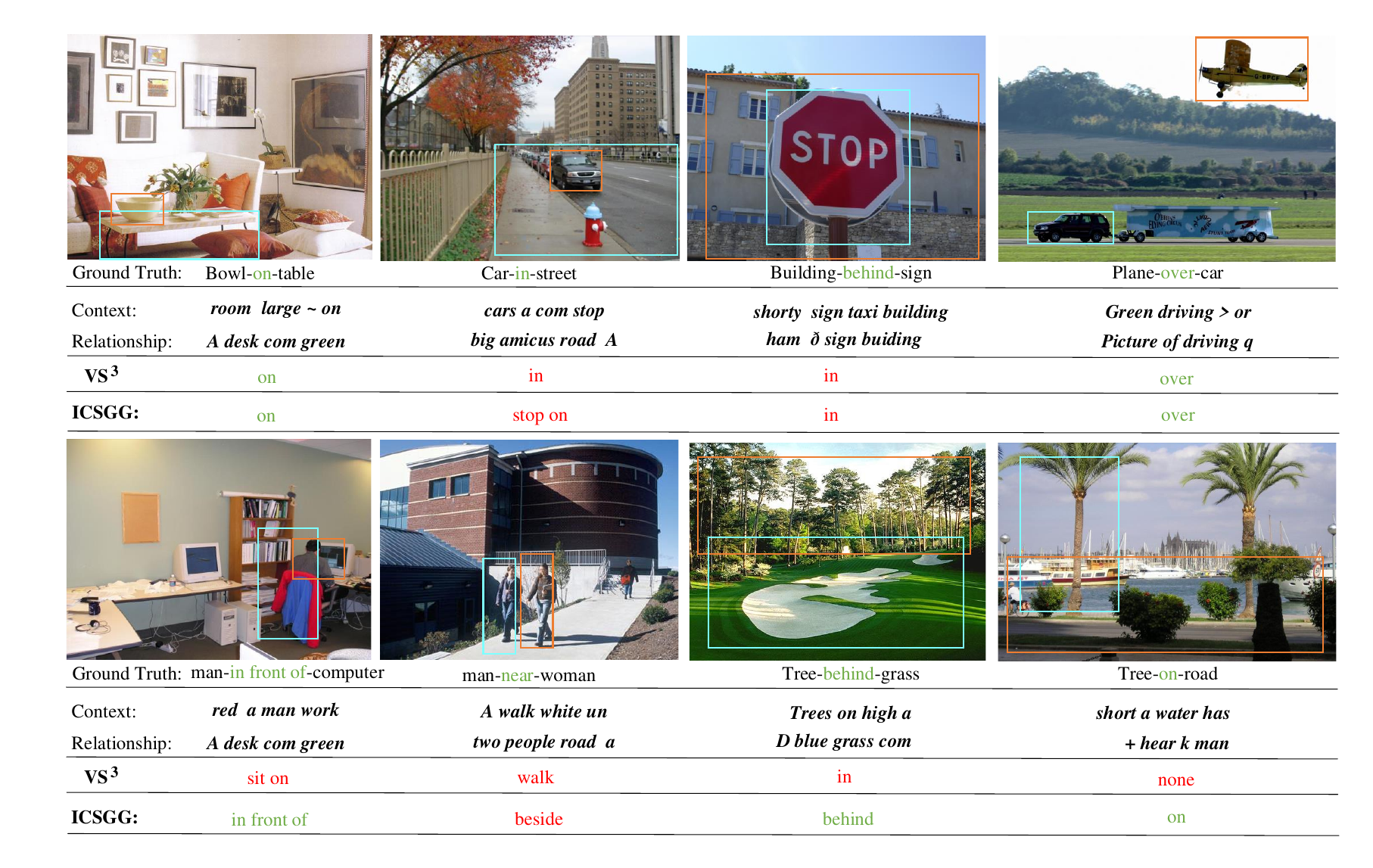} 
 	\caption{ Visualization results of four images from the VG. We   show our transformed textual representations for the context, relationship and the comparison results on the task of \textsc{PredCls} with VS$^3$ \cite{zhang2023learning}.   }
 	\label{fig:vis}
 \end{figure*}

 \subsubsection{Exemplar selection strategies} To validate our proposed knowledge-based retrieval strategy, we compare it against two random selection approaches: (1) randomly selecting $K$ prompts from the prompt set and (2) randomly choosing one sample from the corresponding exemplar set, denoted as $\mathrm{w/o\mbox{-}kap}$ and $\mathrm{w/o\mbox{-}tpo}$, respectively. Additionally, we examine the random ordering strategy $\mathrm{w/o\mbox{-}aso}$ as an alternative to our ascending similarity order. The results indicate that all random strategies exhibit relative performance declines, with the most substantial decrease observed when randomly selecting prompts, which incurs an average decrease of about $2$ points. This affirms the effectiveness of our proposed knowledge-based retrieval strategy for exemplar selection. Moreover, this conclusion aligns with the findings in \cite{min2022rethinking}. 

  \begin{table}[t]
 	\centering
  \caption{ Ablation study of our proposed ICSGG on VG.  }
 	{
 		\begin{tabular}{l|ccc}
 			\toprule
 			\multirow{2}{*}{Methods} & \multicolumn{1}{c}{\textsc{PredCls}} & \multicolumn{1}{c}{\textsc{SGCls}} & \textsc{SGDet}     \\
 			& mR@50/100                   & mR@50/100                & mR@50/100 \\\midrule
 			w/o-kap &   15.3 / 17.1  &  8.5 / ~~9.3    &   4.0 / 5.2        \\
 			w/o-toe &   18.1 / 19.4  &  9.2 / 10.8     &   4.5 / 5.4       \\
 			w/o-aso &   17.5 / 19.0  &  9.0 / 10.4     &   4.2 / 5.3    \\	\midrule
 			w/o-inc &  16.6 / 18.2  &  8.2 / ~~9.5    &   3.6 / 4.8 \\
 			w-1k &  18.1 / 19.6  &  9.2 / 11.0    &   4.6 / 5.5 \\
 		 
 	 	w-ft    &  17.0 / 18.6  &  8.8 / ~~9.6     &   {4.4} / 5.0     \\
    
    \midrule
    w-sc &  17.7 / 19.1  &  8.5 / 10.5    &   4.1 / 5.2 \\
 			w-lc &  18.6 / 20.4  &  9.7 / 11.3    &   {4.9} / {5.8} \\  w-frq  & {18.2} / {20.0} &  {9.8} / {11.1} &  {4.8} / {5.7} \\\midrule
 			{\textbf{ICSGG}}&  {18.6} / {20.3} &  {9.7} / {11.4} &  {4.9} / {5.6}         \\ \bottomrule
 		\end{tabular}
 	}
 	
 	\label{tb:ab}
 \end{table}
 \begin{table}[t]
	\centering
 	\caption{ Ablation study of our proposed ICSGG on Open-Image(v$_6$).  }
		\begin{tabular}{lcccc}
			\toprule
			{Models}& Recall@50    &  wmAP$_{rel}$   &     wmAP$_{phr}$   & score$_{wtd}$ \\  \midrule
		 
			w/o-kap    &  61.71  & 22.37   & 25.29 &   31.41   \\
			w/o-toe    &  64.18  & 25.05   & 27.52  &   33.86      \\
		  w/o-aso    &  63.29  & 24.84   & 26.17  &  33.06      \\\midrule
            
            w/o-inc   &  62.22   & 23.82   & 26.08  &  32.40    \\ 
            w-1k      &  64.63   & 25.20   & 28.10  &    34.25    \\
			w-ft     &   60.17    & 22.49   & 23.47  &   30.42  \\ \midrule
   	
            w-sc       &  63.52  & 23.95   & 26.50  &   32.88   \\ 
            w-lc       &  65.18  & 26.06   & 28.61  &   34.90   \\
            w-frq     & {65.16}   &   {26.10}   & {28.04}   &   {34.71}   \\
			\midrule
 
			{\textbf{ICSGG}}    &  {65.24}   &   {25.97}   & {28.39}   &   {34.79}    \\
			\bottomrule
		\end{tabular} 
	\label{tab:ab_oi} 
\end{table}
\subsubsection{The size of memory buffer}  In this part, we conduct experiments to explore the influence of different buffer sizes, as this parameter may impact the quality of selected samples. By default, we set the buffer size to $2,000$, following the empirical choice in \cite{rebuffi2017icarl}. We further examine two additional buffer size settings: (1) $\mathrm{w/o\mbox{-}inc}$ - no memory buffer, which utilizes Eq.(\ref{eq:l2p}) without the in-context learning strategy for predicting relationships; and (2) $\mathrm{w\mbox{-}1k}$ - setting the buffer size to $1,000$. For the $\mathrm{w/o\mbox{-}inc}$ variant model, we observe significant performance declines, with an average decrease of approximately $2$ points compared to the full  ICSGG. This  demonstrates that a naive prompt alone may not exhibit its superiority in mitigating knowledge forgetting. However, when incorporating more inductive knowledge, such as real samples with ground-truth labels, the more sophisticated in-context prompt can effectively resist knowledge forgetting.

On the other hand, when setting the memory buffer to a smaller size, we observe a slight decrease in performance across all metrics. This phenomenon is likely due to the larger buffer size improving the quality of retrieved samples, which, in turn, benefits the quality of the in-context prompt.

\subsubsection{Finetuning the language model} 
During training, we refrain from updating the language model by default, based on the assumption that fine-tuning may alter the preserved knowledge in the pretrained language model. To validate this assumption, we conduct an experiment involving the finetuning of the pretrained language model, denoted as $\mathrm{w\mbox{-}ft}$. During the finetuning process, we adopt a smaller learning rate for the language model, as done in \cite{he2022towards}. The results indicate that updating the parameters of the language model does not yield any improvement; instead, a noticeable decline is observed across all metrics. This affirms our conjecture that updating the language model would compromise the preserved knowledge and, consequently, have a detrimental effect on the model's generality.
\subsubsection{The length of prompts} Intuitively, the length of input tokens can influence the output of a language model. In our default setting, we configure the length of context, relationship, and object as $4$, $2$, and $2$, respectively. In this experiment, we test two alternative lengths: (1) $\mathrm{w\mbox{-}sc}$, indicating a small length, i.e., $2$, $1$, $1$; and (2) $\mathrm{w \mbox{-} lc}$, indicating a large length, i.e., $8$, $4$, $4$, for context, relationship, and object, respectively.
The results show that the smaller length leads to a slight performance decrease, primarily because shorter prompts result in inferior representations for visual features, capturing less visual content. However, when enlarging the length, the performance increment is limited,  and it incurs additional training time costs.

\subsubsection{The task splitting strategy}
 By default, we employ a random splitting strategy to obtain our subtasks. However, in practical scenarios, later arriving predicates are often rarer. To address this, we investigate an alternative splitting strategy based on predicate frequency, ordering from head to tail. The results under this frequency-based splitting are presented in Table \ref{tab:ab_oi}, denoted as $\mathrm{w\mbox{-}frq}$. From the results, it is observed that the splitting strategy does not exert a significant influence on the overall results, although there is approximately a $0.3$ decrease compared to ICSGG. This suggests that our in-context prompt learning exhibits few-shot learning capability, as the model can handle a severe data-starvation problem, particularly with the frequency-based splitting, especially for tail predicates.

 \subsection{Qualitative Results}
Figure \ref{fig:vis} showcases four qualitative results comparing ICSGG and the state-of-the-art model VS$_3$  \cite{dong2022stacked} on the \textsc{PredCls} task after five training stages. Additionally, we present the embedding tokens of the context and relationship using our $\mathrm{VL2T}$. To present our transformed textual tokens, we choose the closest embedding from the vocabulary of GPT-$2$, following  \cite{mokady2021clipcap}.
From the results, it is evident that although our textual tokens do not construct a conventional sentence, certain key content is preserved. For instance, in the first figure, the context presentations contain terms such as "room" and "on". Remarkably, in the second example, our model predicts the entirely novel predicate "stop on", which is more informative than the ground truth "on".
\section{Conclusion}
In this paper, we propose a novel and practical task for scene graph generation, dubbed lifelong scene graph generation, aiming to learn to predict relationship in a incremental fashion. Towards LSGG, we have to solve two intrinsic challenges: extreme imbalanced data distribution and forgetting. To sovle those problems, we proposed to use textual embeddings to present visual content so that we could leverage the pretrained language model to infer relationships by designing in-context based prompts. Extensive experiments show that our proposed ICSGG is   much  better than other SOTA SGG models. \textbf{Limitations:} we do not jointly train the object detection network in this paper by designing a one-step framework. In the future work, we will explore   a  one-step model for LSGG in an end-to-end manner to predict relationship.
	
	{ 
		\bibliographystyle{IEEEtran}
		\bibliography{egbib}
	}

\end{document}